\documentclass[10pt,twocolumn,letterpaper]{article}

\usepackage[final]{iccv}         

\usepackage{graphicx}
\usepackage{amsmath}
\usepackage{amssymb}
\usepackage{booktabs}
\usepackage{enumitem}
\usepackage{tabularx}
\usepackage{makecell}
\usepackage{enumerate}
\usepackage[pagebackref,breaklinks,colorlinks]{hyperref}

\usepackage[capitalize]{cleveref}
\crefname{section}{Sec.}{Secs.}
\Crefname{section}{Section}{Sections}
\Crefname{table}{Table}{Tables}
\crefname{table}{Tab.}{Tabs.}

\def\paperID{4} % *** Enter the WACV Paper ID here
\def\confName{ICCV}
\def\confYear{2025}

\begin{document}
\renewcommand{\thefootnote}{\fnsymbol{footnote}}
\title{A Survey on Vision-Language-Action Models for Autonomous Driving}

\author{Sicong Jiang$^1$$^\ast$, Zilin Huang$^4$$^\ast$, Kangan Qian$^2$$^\ast$, Ziang Luo$^2$, Tianze Zhu$^2$, Yang Zhong$^3$, 
Yihong Tang$^1$, \\Menglin Kong$^1$, Yunlong Wang$^2$, Siwen Jiao$^3$, Hao Ye$^3$, Zihao Sheng$^4$, Xin Zhao$^2$, Tuopu Wen$^2$, \\Zheng Fu$^2$, Sikai Chen$^4$, Kun Jiang$^2$$^,$$^6$,  Diange Yang$^2$$^,$$^6$, Seongjin Choi$^5$, Lijun Sun$^1$\\
$^1$ McGill University, Canada \\$^2$ Tsinghua University, China \\ $^3$ Xiaomi Corporation \\$^4$ University of Wisconsin–Madison, USA \\$^5$ University of Minnesota–Twin Cities, USA\\$^6$ State Key Laboratory of Intelligent Green Vehicle and Mobility, Tsinghua University, China\\
{\tt\small sicong.jiang@mail.mcgill.ca, lijun.sun@mcgill.ca}
}
\maketitle
%%%%%%%%% ABSTRACT
\begin{abstract}
% The rapid progress of multimodal large language models (MLLM) has paved the way for Vision-Language-Action (VLA) paradigms, which integrate visual perception, natural language understanding, and control within a single policy. Researchers in autonomous driving are actively adapting these methods to the vehicle domain. Such models promise autonomous vehicles that can interpret high-level instructions, reason about complex traffic scenes, and make their own decisions. However, the literature remains fragmented and is rapidly expanding. This paper provides the first comprehensive survey on VLA for Autonomous Driving (VLA4AD). In this survey, we (i) trace its evolution from explainer to reasoning-centric models, (ii) formalize the architectural building blocks shared across recent work, and (iii) organize over twenty representative models according to VLA’s progress in the autonomous driving domain. We further consolidate the landscape of multimodal datasets and benchmarks, highlighting how new evaluation protocols jointly measure driving safety, instruction fidelity, and explanation quality. Finally, we distill open challenges in robustness, real-time efficiency, and safety verification, and chart future directions toward foundation-scale driving models and standardized “traffic language” protocols. This survey provides a concise yet comprehensive reference to accelerate research on interpretable, instruction-following, and socially compliant autonomous driving.

The rapid progress of multimodal large language models (MLLM) has paved the way for Vision-Language-Action (VLA) paradigms, which integrate visual perception, natural language understanding, and control within a single policy. Researchers in autonomous driving are actively adapting these methods to the vehicle domain. Such models promise autonomous vehicles that can interpret high-level instructions, reason about complex traffic scenes, and make their own decisions. However, the literature remains fragmented and is rapidly expanding. This survey offers the first comprehensive overview of VLA for Autonomous Driving (VLA4AD). We (i) formalize the architectural building blocks shared across recent work, (ii) trace the evolution from early explainer to reasoning-centric VLA models, and (iii) compare over 20 representative models according to VLA’s progress in the autonomous driving domain. We also consolidate existing datasets and benchmarks, highlighting protocols that jointly measure driving safety, accuracy, and explanation quality. Finally, we detail open challenges—robustness, real-time efficiency, and formal verification—and outline future directions of VLA4AD. This survey provides a concise yet complete reference for advancing interpretable socially aligned autonomous vehicles. Github repo is available at \href{https://github.com/JohnsonJiang1996/Awesome-VLA4AD}{JohnsonJiang/Awesome-VLA4AD}.
\end{abstract}
{\renewcommand{\thefootnote}{}%
\footnotetext{$^\ast$Equal contribution.}%
}

%%%%%%%%% BODY TEXT
\section{Introduction}
\label{sec:intro}

Autonomous vehicles must simultaneously \emph{perceive} complex 3D scenes, \emph{understand} traffic context, and \emph{act} safely in real time. Classic autonomous driving (AD) stacks achieve this through a hand-engineered cascade of perception, prediction, planning, and control modules. While decades of research have made such pipelines reliable under common conditions, they remain brittle at module boundaries and struggle with long-tail corner cases - scenarios that demand high-level reasoning or nuanced human interaction. 

Progress in foundation models, such as vision-language models (VLMs) and large language models (LLMs), has introduced strong semantic priors into driving perception. By aligning pixels with text, these models can explain scenes, answer questions, or retrieve contextual information that traditional detectors may miss \cite{radford2021learning,liu2023visual,jia2023adriveri,sima2024drivelm,qian2025fasionad++}. Early adaptations have improved generalization to rare objects and provided human-readable explanations, e.g., describing an ambulance’s trajectory or justifying a red-light stop. However, VLM-augmented stacks remain \emph{passive}: they reason \emph{about} the scene but do not decide \emph{what to do}. Their language output is loosely coupled to low-level control and may hallucinate hazards or misinterpret colloquial instructions. In short, while VLMs enhance interpretability, they leave the action gap unresolved.

\begin{figure*}[htp]
    \centering
    \vspace{-10pt}
    \includegraphics[width=1\linewidth]{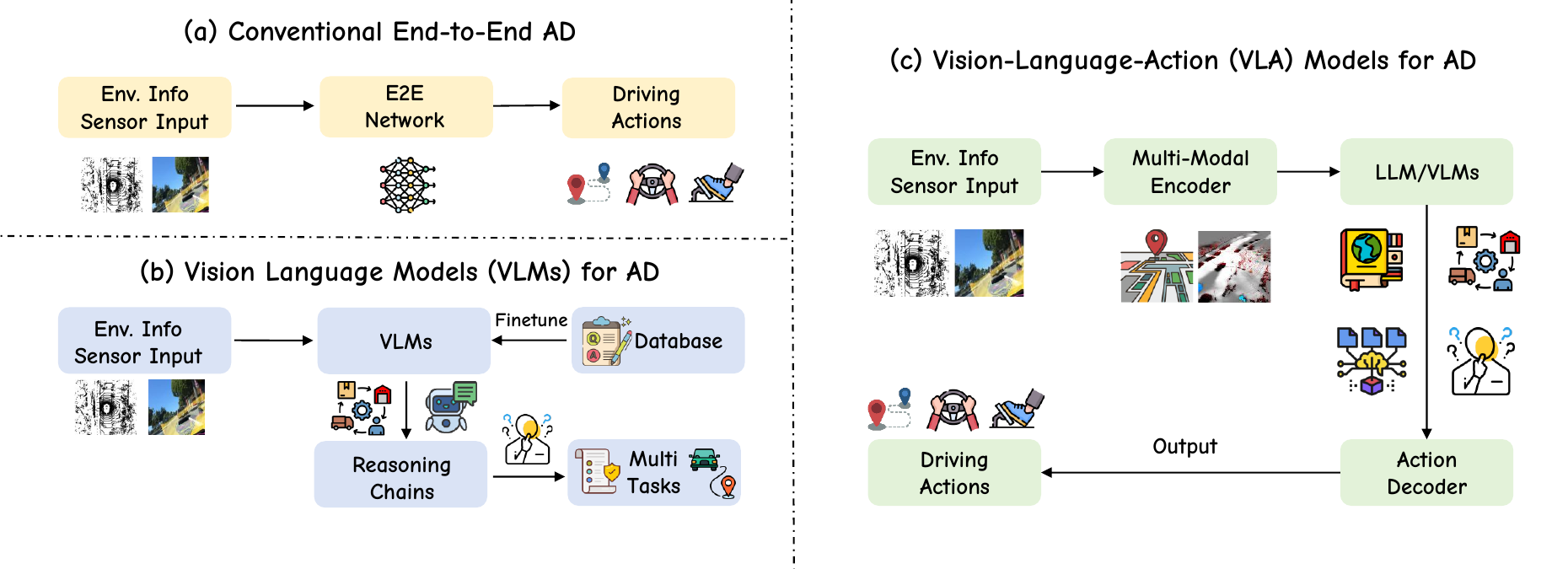}
    \vspace{-20pt}
    \caption{Comparisons of autonomous driving paradigms. (a) End-to-end driving offers direct perception-to-control mapping but lacks interpretability and generalization. (b) VLM4AD introduces natural language reasoning and explainability, yet remains perception-centric. (c) VLA4AD integrates perception, reasoning, and action, enabling interpretable and robust closed-loop control.}
 \vspace{-5pt}
    \label{fig1}
\end{figure*}

Recent work has thus proposed a more integrated paradigm: VLA models that fuse camera streams, natural language instructions, and low-level actuation within a single policy \cite{chi2024covla,jiang2025diffvla,zhou2025opendrivevla}. By conditioning the controller on language tokens, these systems can (i) follow free-form commands such as “yield to the ambulance” \cite{renz2025carllava}, (ii) verbalize their internal rationale for post-hoc verification \cite{chi2024covla}, and (iii) leverage commonsense priors from internet-scale corpora to extrapolate in rare or unforeseen situations \cite{chi2025impromptu}. Early prototypes have already demonstrated improved safety and instruction fidelity in simulation and closed-loop tests, foreshadowing a new research frontier we term \emph{VLA for Autonomous Driving (VLA4AD)}.

Several converging trends underscore the timeliness of this emerging research frontier. First, petabyte-scale multi-sensor logs such as nuScenes \cite{caesar2020nuscenes} and Impromptu VLA \cite{chi2025impromptu} provide rich multimodal supervision. Second, foundation LLMs can now be efficiently adapted through low-rank updates, while token-reduction designs such as TS-VLM \cite{chen2025tsvlm} reduce online computational cost by an order of magnitude \cite{zhou2025dynrslvlm}. Third, synthetic corpora like SimLingo \cite{renz2025carllava} and interactive datasets such as NuInteract \cite{zhao2025drivemonkey} enable researchers to stress-test language-conditioned behaviors well before real-world deployment. Together, these developments have sparked a surge of VLA architectures, ranging from explanatory overlays to reasoning-centric agents with chain-of-thought (CoT) memory.

Although several surveys now cover the application of LLMs and VLMs to autonomous driving \cite{cui2024survey,zhou2024vision,cui2025chain,gao2024survey}, none have yet addressed the rapidly emerging VLA paradigm in AD. To close this gap and consolidate this rapidly expanding body of work, we present the first comprehensive survey of VLA4AD. We first clarify key terminology and relate VLA to traditional end-to-end driving. Then, we distill common architectural patterns and catalogue over twenty representative models along with the datasets that support them. Also, we compare training paradigms and summarize evaluation protocols that jointly assess control performance and language fidelity. Finally, we outline open challenges and chart promising future directions.We also highlight the need for standardized benchmarks and open-source toolkits to foster reproducibility and accelerate cross-model comparisons. Our goal is to provide a coherent and forward-looking reference on how vision, language, and action are converging to shape the next generation of transparent, instruction-following, and socially compliant autonomous vehicles. 
 
\section{Development of Autonomous Driving}
\label{sec:DevelopmentofAD}
The technical arc of AD can be traced through four
main paradigms: \emph{modular stacks}, \emph{end-to-end learning}, \emph{VLM4AD}, and
the recent \emph{VLA4AD} wave.

\subsection{Classical Modular Pipelines}

The first wave of AD systems-epitomized by the DARPA Urban Challenge vehicles-explicitly factorized the driving task into distinct modules: perception\cite{hussain2023yolo, liu2023bevfusion, zhang2022beverse}, prediction\cite{altche2017lstm, huang2022survey}, planning\cite{schwarting2018planning, hu2023planning}, and control \cite{urmson2008darpa,paden2016survey,gonzalez2015review,kendall2019learning,chen2021data,jiang2024communication}. Hand-crafted algorithms processed LiDAR, radar, and GPS data: traditional vision detectors identified objects, finite-state machines or graph-based search generated paths, and PID or MPC controllers executed the final commands \cite{lefevre2014survey,ismail2024narrate}. This architecture was widely adopted in industry due to its modularity-each component could be engineered, tested, and improved in isolation. However, such strict decoupling leads to information fragmentation: upstream errors propagate without correction, and misaligned objectives across modules hinder end-to-end optimization \cite{liang2020pnpnet,luo2018fast,sadat2020perceive}.
% While co-training promotes hierarchical feature sharing and reduces computational cost, it also introduces the challenge of negative transfer-inter, i.e., task interference that can degrade overall performance \cite{crawshaw2020multi,liu2023bevfusion}.
% To mitigate these issues, subsequent systems embraced multi-task learning (MTL), where a shared backbone feeds multiple task-specific heads for detection, segmentation, motion prediction, and planning. Notable examples include 
% %Transfuser \cite{chitta2022transfuser}, 
% BEVerse \cite{zhang2022beverse}, and unified architectures proposed in \cite{chen2022learning,zeng2019end}. Similar approaches underpin industrial stacks at companies such as Mobileye, Tesla, and NVIDIA. 

% While interpretable and verifiable, these modular systems required extensive engineering and tuning, and struggled to generalize under complex or rare driving scenarios.

\subsection{End-to-End Autonomous Driving}
The modular design of autonomous driving systems suffers from error propagation and information loss across the perception, prediction, and planning modules. Consequently, research has shifted towards more integrated, end-to-end approaches. As illustrated in Fig.~\ref{fig1} (a), end-to-end(E2E) driving policies map raw sensor streams directly to control commands, bypassing hand-crafted modular pipelines \cite{chen2022learning, zeng2019end, hawke2020urban, chitta2021neat, renz2022plant, chitta2022transfuser, shao2023safety, jia2023think, shao2023reasonnet, feng2024road, ye2023fusionad, xu2024m2da}. 

E2E driving fundamentally operates as a Vision-to-Action (VA) system, where visual input can be from cameras or LiDAR, and the action output is typically represented as future trajectories or control signals. However, directly mapping raw sensory input to driving actions presents significant challenges due to the sparsity of planning-level data and the vast, unstructured solution space inherent in neural networks\cite{chen2024end}. To mitigate these issues, early E2E approaches introduced intermediate supervision through integrated perception and prediction tasks\cite{chitta2022transfuser, shao2023safety}. Specifically, these methods integrate perception, prediction, and planning modules within a unified framework, primarily facilitating feature-level information flow across modules. UniAD~\cite{hu2023uniad} primarily relies on rasterized representations (e.g., semantic maps, occupancy maps, flow maps, and costmaps), which is computationally intensive. In contrast, the proposed VAD~\cite{jiang2023vad, ma2025leapvad} adopts a fully vectorized scene representation for end-to-end planning. VAD leverages instance-level structural information as both constraints and guidance for planning, achieving promising performance with higher efficiency. PolarPoint-BEV~\cite{feng2024polarpoint} further refines the BEV representation by employing a polar point encoding. This incorporates a distance-based importance weighting prior, enabling the model to focus more effectively on critical objects at varying ranges during driving. 

To model the interactions between the ego vehicle and other traffic participants, GenAD~\cite{zheng2024genad} and PPAD\cite{chen2024ppad} leverages instance-level visual features, whereas GraphAD~\cite{zhang2024graphad} represents these features as nodes in a graph. SparseAD~\cite{zhang2024sparsead} and SparseDrive~\cite{sun2024sparsedrive}formulates a fully sparse architecture, achieving greater efficiency and superior planning performance. However, these methods typically rely on constructing computationally expensive BEV features and prevent downstream tasks from learning directly from raw sensor inputs. To mitigate this challenge, PARA-Drive~\cite{weng2024paradrive} and DriveTransformer~\cite{jia2025drivetransformer} introduces a parallel pipeline architecture, enabling a more unified and scalable framework for end-to-end driving systems. This approach explicitly models the relations between perception, prediction, and planning tasks. However, these paradigms also fall short in handling corner cases and struggle with out-of-distribution (OOD) scenarios. For instance, a driving system might fail to generate appropriate trajectories when encountering rare events, such as a vehicle breaking down at an intersection \cite{xu2024vlm}.

Several methods attempt to mitigate these issues. Physics-law enhanced frameworks incorporate prior knowledge \cite{zhou2024enhance}, while others utilize objective functions and planning action priors to refine trajectories \cite{jia2023think, wang2023s, li2024hydra, shen2024utilizing, chen2024vadv2, li2025generalized, liu2025two, yao2025drivesuprim, guo2025ipad}. However, designing such objective functions for trajectory refinement remains both computationally expensive and labor-intensive. Subsequent research aims to reduce annotation burdens for 3D perception tasks through self-supervised or weakly-supervised frameworks, such as \cite{li2024enhancing, guo2024end, li2024does, lu2024activead}.

Furthermore, closed-loop evaluations reveal diminishing returns beyond certain data volumes, coupled with significant performance variance across different scenario types \cite{naumann2025data,zheng2024preliminary}. These findings indicate that pure data scaling alone is insufficient for achieving Level 4+ autonomy.

Overall, end-to-end learning has significantly narrowed the gap between raw sensor inputs and control decisions. Yet, persistent challenges remain, including:
(i)  Brittle semantics: Vulnerability to rare or rapidly evolving scenarios.
(ii) Opaque reasoning: Limited interpretability hindering safety auditing and verification.
(iii) Limited language proficiency: Restricting intuitive human-vehicle interaction.

\subsection{VLMs for Autonomous Driving}
Combining language modalities with driving tasks provides a promising direction to enhance reasoning, interpretability, and generalization in autonomous driving systems. LLMs \cite{touvron2023llama1,touvron2023llama2} and VLMs \cite{radford2021learning, alayrac2022flamingo, liu2023visual} offer a promising remedy by unifying perception and natural language reasoning within a shared embedding space \cite{huang2024vlm}. At the core of this progress is large-scale multimodal pretraining, which equips models with commonsense associations (e.g., siren → yield) that task-specific labels often miss. Consequently, language-conditioned VLM policies exhibit stronger zero-shot generalization across novel objects, weather conditions, and driving norms. Recent work \cite{fu2024drive, wang2024drive, liu2023mtd, wang2023drivemlm, ma2024dolphins, paul2024lego, zhang2024instruct, huang2024drivemm} has begun embedding these models directly into the driving loop, as illustrated in Fig.~\ref{fig1} (b).

Early efforts including GPT-Driver \cite{mao2023gpt}, which demonstrates that frozen VLMs can process multi-view images and textual prompts to generate trajectory plans or low-level control tokens while simultaneously producing human-readable rationales. While effective for commonsense reasoning and corner case understanding, integrating large foundation models into driving systems presents several drawbacks: poor spatial awareness \cite{guo2024drivemllm}, ambiguous numerical outputs \cite{jiang2024senna}, and elevated planning latency \cite{tian2024drivevlm}. Furthermore, hallucination effects prevalent in LLMs/VLMs \cite{xie2025vlms} expose driving systems to potentially unsafe control signals.

Follow-up research addresses these limitations through several key directions:
(1) Spatial Understanding Enhancement: TOKEN~\cite{tian2024tokenize} and WKAD~\cite{zhai2025world} use object-centric token representations, while BEVDriver~\cite{winter2025bevdriver} integrates BEV features with language for 3D spatial queries and multimodal future predictions. Sce2DriveX~\cite{zhao2025sce2drivex} proposes spatial relationship graphs to model interactions between ego vehicle and traffic participants, and MPDrive~\cite{zhang2025mpdrive} leverages visual prompting to strengthen spatial reasoning.
(2) Latency Reduction: Dual-system architectures \cite{tian2024drivevlm, mei2024leapad, ma2025leapvad, jiang2024senna, chen2024asynchronous, doll2024dualad, ding2024hint, miyaoka2024chatmpc, long2024vlmmpc, qian2025fasionad++} employ VLMs as intermediate modules providing feedback or auxiliary signals to end-to-end systems. Knowledge distillation approaches \cite{han2025dmedriver, pan2024vlp, xu2024vlm, liu2025vlm} transfer VLM capabilities to traditional systems offline.
(3) Hallucination Mitigation: In-context learning methods like Dilu~\cite{wen2023dilu} and its extension~\cite{jiang2024koma} utilize memory banks to store critical driving information. ReasonPlan~\cite{liu2025reasonplan} generate step-by-step decision justifications. AgentDriver~\cite{mao2023agentdriver} and AgentThink~\cite{qian2025agentthink} implement tool-augmented chain-of-thought prompting to enhance reasoning capabilities.

Despite these advances, current methods remain predominantly perception-centric: their generated plans lack tight integration with closed-loop control\cite{shao2024lmdrive}, and their explanatory outputs provide no formal safety guarantees\cite{zhou2024enhance}. Besides, how to align VLM's outputs with the action space is also a challenge.

% At the core of this progress is large-scale multimodal pretraining, which equips models with commonsense associations (e.g., siren → yield) that task-specific labels often miss. As a result, language-conditioned VLM policies exhibit stronger zero-shot generalization across novel objects, weather, and driving norms. marking a pivotal step toward human-aligned driving.

\subsection{From VLM to VLA for Autonomous Driving}
% \label{subsec:vlarise}

Inspired by recent progress in the embodied intelligence field \cite{black2024pi0, kim2024openvla, liu2024robomamba}, aligning vision, language, and action within a unified framework has become a growing trend in autonomous driving.
As shown in Fig.~\ref{fig1} (c), VLA addresses the aforementioned gap by incorporating an explicit action head, thereby unifying perception, reasoning, and control within a single policy \cite{ma2024surveyvla, sapkota2025vla, zhou2024vision}. VLA policies are driven by three core demands in real-world driving: (i) robust reasoning in rare and long-tail scenarios \cite{jia2024bench2drive, neurips2024drivingdojo}; (ii) noise-tolerant control under dynamic and partially occluded conditions; and (iii) the ability to interpret spontaneous, high-level language commands (e.g., “overtake the truck”) \cite{pan2024vlp, xu2024drivegpt4}. By leveraging foundation models pretrained on internet-scale visual and linguistic data \cite{oquab2023dinov2}, VLA models demonstrate strong generalization across domains and benchmarks \cite{nuscenes2019, wilson2023argoverse2, waymo2024open}.
Concretely, modern VLA models can: (i) ground free-form instructions within ego-centric visual contexts and generate corresponding trajectories \cite{yuan2024rag,fu2025orion}; (ii) produce Chain-of-Thought (CoT) justifications, as seen in DriveCoT and CoT-VLA \cite{wang2024driveCoT,cui2025chain}, to enhance interpretability; and (iii) move beyond direct control tokens to incorporate advanced planning modules, including diffusion-based heads \cite{jiang2025diffvla}, hierarchical CoT-based planners \cite{fu2025orion}, and hybrid discrete–continuous control strategies.

Recent exemplar systems highlight the breadth of VLA capabilities\cite{chi2024covla,fu2025orion,zhang2025safeauto,jiang2025diffvla}, exemplifing the new VLA4AD paradigm: they jointly reason over vision, language, and action, combining textual and trajectory outputs, long-horizon memory, symbolic safety checks, and multi-modal diffusion planning. These advances represent a decisive shift from perception-centric VLM pipelines toward action-aware, explainable, and instruction-following multimodal agents—paving the way for safer, more generalizable and human-aligned autonomous driving.

\section{Architecture Paradigm of VLA4AD}
\begin{figure*}
    \centering
    \includegraphics[width=1\linewidth]{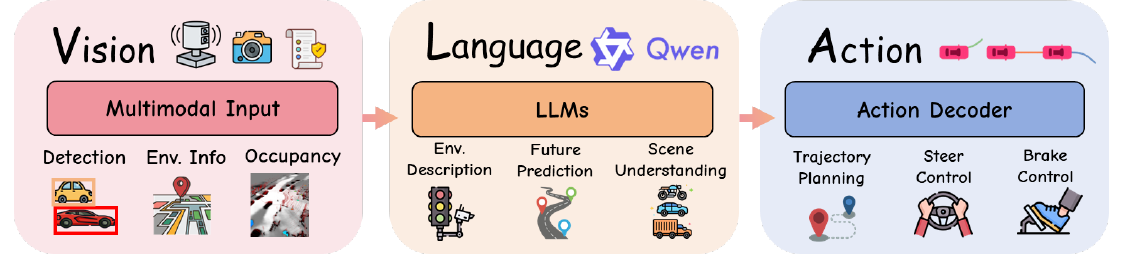}
    \caption{Overview of the VLA4AD Architecture.}
 \vspace{-5pt}
    \label{fig2}
\end{figure*}
% The development of VLA models for autonomous driving has been driven by the limitations of earlier paradigms---modular pipelines, multi-task learning frameworks, and even end-to-end policies---which often lack transparency, semantic alignment, or flexibility. Converging advances in VLMs and LLMs provide a new foundation for unifying perception, reasoning, and control. 

This section presents the basic architecture of VLA4AD, consolidating its multimodal interfaces, core components, and outputs, as illustrated in Fig.~\ref{fig2}.

\subsection{Multimodal Inputs and Language Commands}

VLA4AD rely on rich multimodal sensor streams and linguistic inputs to capture both the external environment and the driver's high-level intent.

% \paragraph{Visual Data.} Human can overcome large amount of driving scenes with their eyes. visual information plays vital importance in the autonomous driving tasks. This is a primary modality, typically consisting of video streams from one or more cameras mounted on the vehicle. While front-facing cameras are common, comprehensive scene understanding often requires surround-view camera systems \cite{waymo2024open}. Raw images may be processed directly or transformed into intermediate representations such as BEV maps, which simplify spatial reasoning \cite{researchgate2025talk2bev}. The resolution of input images is a key design parameter, involving trade-offs between visual detail and computational efficiency \cite{zhou2025dynrslvlm}.
\paragraph{Visual Data.}
Humans rely heavily on visual input to navigate complex driving environments, so do autonomous systems. In early approaches, single front-facing monocular cameras were the standard visual modality\cite{yebes2015visual, marcu2024lingoqa, kim2018bddx}. Over time, to improve spatial coverage and safety, systems evolved to include stereo cameras, multi-camera setups, and eventually full surround-view systems\cite{nuscenes2019, waymo2024open}. This richer visual input enables more robust scene understanding and multi-object reasoning. Raw images can be processed directly or transformed into structured intermediate representations, such as bird’s-eye-view (BEV) maps that facilitate spatial reasoning \cite{researchgate2025talk2bev, xu2025chatbev}. Recent work further explores the trade-off between input resolution and model efficiency, dynamically adjusting granularity for real-time or long-tail cases \cite{zhou2025dynrslvlm}.

\paragraph{Other Sensor Data.}
Beyond vision, autonomous vehicles have increasingly leveraged diverse sensor modalities to complement and ground perception to enhance spatial capabilities. Initial systems integrated LiDAR for precise 3D structure, later combining it with RADAR for velocity estimation and IMUs for motion tracking. GPS modules provide global localization \cite{nuscenes2019, wilson2023argoverse2}. The field has also seen increasing attention to proprioceptive data, such as steering angle, throttle, and acceleration, particularly for behavior prediction and closed-loop control\cite{chitta2022transfuser, winter2025bevdriver, yang2025lidar}. This progression—from geometry to dynamics—has driven research into more sophisticated sensor fusion frameworks \cite{li2025pointvla, broekens2023fine, wei2024occllama, chen2025insight}, aiming to create a unified spatial-temporal representation of the environment.

\paragraph{Language Inputs.}
Natural language inputs—such as commands, queries, and structured descriptions—have become increasingly important in VLA4AD. Early research focused on direct navigation commands (e.g., “Turn left at the next intersection,” “Stop behind the red car”) to enable basic instruction following \cite{pan2024vlp, renz2025carllava}. As systems matured, environmental queries emerged, allowing users or agents to ask questions like “Is it safe to change lanes now?” or “What is the speed limit here?” \cite{nie2024reason2drive, ishaq2025drivelmm}, enabling interactive situational awareness.
Further advancements introduced task-level linguistic specifications, such as interpreting traffic rules, parsing high-level goals, or understanding map-based constraints expressed in natural language \cite{gao2025langcoop}. More recent efforts have pushed toward multi-turn dialogs, reasoning chains (e.g., Chain-of-Thought prompting)\cite{tian2024drivevlm, hwang2024emma}, and tool-augmented language interfaces\cite{mao2023agentdriver, hou2025driveagent, qian2025agentthink}, which support richer forms of reasoning and alignment with human decision-making processes.

Finally, recent work has also started incorporating spoken language as a more natural and embodied input modality, bridging perception and interaction via speech-driven interfaces \cite{yuan2024rag,zhou2025opendrivevla}. This progression from static instructions to dialog-driven, multi-step reasoning reflects a broader trend: using language not just to command the vehicle, but to enable interpretable and collaborative autonomy.

\subsection{Core Architectural Modules}

The fundamental architecture of a VLA4AD integrates visual perception, language understanding, and action generation in a cohesive pipeline.

\paragraph{Vision Encoder.} Raw imagery and sensor data are converted to latent representations using large self-supervised backbones such as DINOv2 \cite{oquab2023dinov2}, ConvNeXt-V2 \cite{woo2023convnext}, or CLIP \cite{radford2021learning}. Many VLA systems employ BEV projection \cite{tian2024drivevlm}, and others incorporate 3D priors via point-cloud encoders (e.g., PointVLA \cite{li2025pointvla}) or voxel modules (3D-VLA \cite{zhen20243d}). Multi-scale fusion using language-derived keys improves grounding at fine spatial levels \cite{zhang2024minidriveiclr,qu2025spatialvla}. 

\paragraph{Language Processor.} Natural language is processed using pretrained decoders such as LLaMA2 \cite{touvron2023llama} or GPT-style transformers \cite{brown2020language}. Instruction-tuned variants (e.g., Visual Instruction Tuning \cite{liu2023visual}) and retrieval-augmented prompting (RAG-Driver~\cite{yuan2024rag}) inject domain knowledge. Lightweight fine-tuning strategies such as LoRA \cite{hu2022lora} enable efficient adaptation.

\paragraph{Action Decoder.} Downstream control is emitted via (i) Autoregressive tokenizers where discrete actions or trajectory way‑points are predicted sequentially \cite{pertsch2025fast,kim2025finetuningvla,hung2025nora, zhou2025autovla}, (ii) Diffusion heads that sample continuous controls conditioned on fused embeddings (DiffVLA \cite{jiang2025diffvla}; Diffusion‑VLA \cite{wen2024diffusion}), or (iii) Flow‑matching / policy gradient experts used by GRPO \cite{shao2024deepseekmath} or DPO\cite{rafailov2023direct} fine‑tuning pipelines \cite{xiong2024autoregressive,li2025recogdrivereinforcedcognitiveframework}. Hierarchical controllers (e.g., ORION \cite{fu2025orion}) let a language planner dispatch sub‑goal sketches to a separate low‑level PID or MPC stack.

% \subsection{Driving Outputs}
% The output of a VLA model reflects its operational goal:

% \paragraph{Low-Level Actions.} Many models directly predict steering, throttle, and brake commands, either as continuous control signals or as discrete control tokens \cite{fu2025orion,zhou2025opendrivevla,yang2025drivemoe,pertsch2025fast}.

% \paragraph{Trajectory Planning.} A more common formulation involves predicting future trajectories or waypoints, typically represented in BEV or ego-centric coordinates \cite{pan2024vlp,jiang2025diffvla,bartoccioni2025vavim}, which are then executed by MPC-based controllers \cite{zhang2024analysis}.

% Together, these outputs reflect the growing ambition of VLA systems: not just to control a vehicle, but to act explainably, robustly, and contextually within complex traffic environments. In summary, VLA4AD  takes multimodal sensor data and a language query or command as input, and produces both driving decisions (e.g., control signals or planned trajectories) and, in some cases, accompanying language-based explanations.

\subsection{Driving Outputs}
The output modality of a VLA model reflects its level of abstraction and operational goal. Over time, output formats have evolved from low-level control commands to higher-level spatial reasoning and skill-conditioned actions.

\paragraph{Low-Level Actions.}
Early VLA4AD systems typically focused on directly predicting raw control signals such as steering angles, throttle, and braking. These actions are often modeled either as continuous outputs or as discrete action tokens, suitable for integration with PID or end-to-end control pipelines \cite{fu2025orion,zhou2025opendrivevla,yang2025drivemoe,pertsch2025fast}. While this formulation allows for fine-grained control, it is often sensitive to small perception errors and lacks long-horizon planning capacity.

\paragraph{Trajectory Planning.}
Subsequent research has shifted towards trajectory- or waypoint-level predictions, which offer a more stable and interpretable intermediate representation. These trajectories, often expressed in BEV or ego-centric coordinates, can be flexibly executed via model predictive control (MPC) or other downstream planners \cite{pan2024vlp,hu2023uniad,jiang2023vad,jiang2025diffvla,bartoccioni2025vavim,zhang2024analysis}. This formulation allows VLA models to reason over longer time horizons and integrate multimodal context more effectively.

% \paragraph{Skill-Level Decisions.}
% More recent approaches explore skill-conditioned or behavior-level outputs, which lie between low-level control and full trajectory specification. These models predict high-level driving intents (e.g., “yield,” “merge,” “lane change”) or discrete maneuver classes, enabling hierarchical planning and improving sample efficiency in learning \cite{wu2024driveskill,xu2025skillplanner}. This abstraction helps bridge the gap between language commands and actionable motion plans.

Together, these output formats illustrate the evolving ambition of VLA4AD systems: not only to drive, but to do so robustly, explainably, and contextually. In summary, a typical VLA4AD model takes multimodal sensor data and natural language input as context, and produces both driving decisions (at various abstraction levels) and, in some cases, language-grounded explanations.

% \section{Key VLA4AD Models}
\section{Progress of VLA4AD Paradigm}

\label{sec:vla_progress_stage}
\begin{figure*}
    \centering
    \vspace{-10pt}
    \includegraphics[width=1\linewidth]{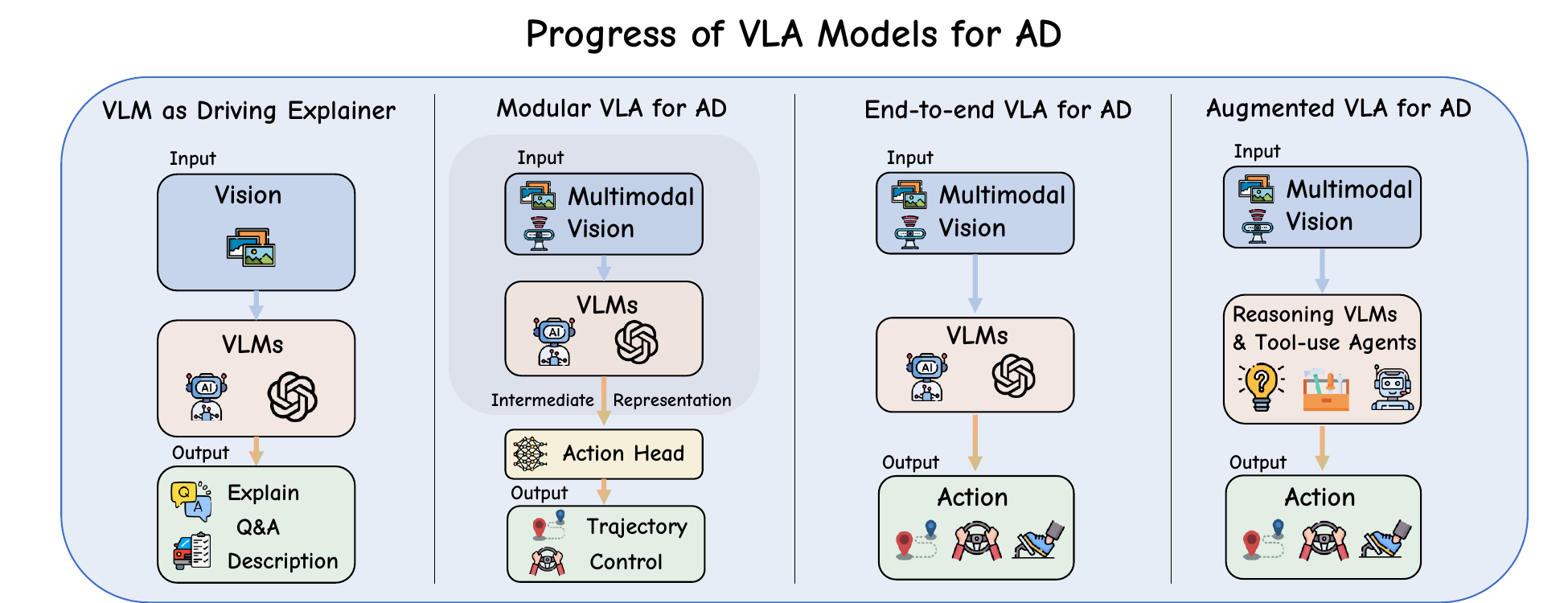}
    \vspace{-15pt}
    \caption{Evolution of VLA models for AD. From left to right:
(1) VLM-as-explainer: a frozen LLM narrates the driving scene but produces no control.
(2) Modular VLA: language is transformed into an intermediate representation that an action head converts into trajectory or low-level control.
(3) End-to-end VLA: a single multimodal pipeline maps sensor input directly to actions.
(4) Augmented VLA: tool-using or CoT VLMs add long-horizon reasoning while retaining the end-to-end control pathway.}
 \vspace{-5pt}
    \label{fig3}
\end{figure*}

Research on the VLA4AD has moved in discernible waves, each propelled by the limitations of its predecessor and by the arrival of new cross-modal pre-training techniques. As shown in Fig. \ref{fig3}, in what follows, we trace four successive stages: \emph{Explanatory Language Models}, \emph{Modular VLA4AD}, \emph{End-to-end VLA4AD}, and \emph{Reasoning-centric VLA4AD}. 

Table \ref{tab:vla4ad_models} summarizes representative VLA4AD models from 2023–2025, highlighting their input modalities, how they incorporate language, the form of action output, the data or environment used for evaluation, and their core contributions.

% In what follows, we trace four successive stages—\emph{explanatory overlay}, \emph{modular planning}, \emph{end-to-end unification}, and \emph{reasoning-centric interaction}—highlighting how the representative systems inside each stage both inherit and transcend the design logic of the one before.  Throughout the discussion.

\subsection{Pre-VLA: Language Model as Explainer}

% The first deployments of large language models in autonomous driving treated language chiefly as a window into the perception stack.  A typical pipeline coupled a frozen vision encoder such as CLIP with an instruction-tuned decoder and asked it to \emph{describe} what the cameras saw.  \textbf{DriveGPT-4}~\cite{xu2024drivegpt4} and \textbf{CoVLA-Agent}~\cite{chi2024covla} exemplify the pattern: single-frame images are mapped to fluent captions or symbolic manoeuvre tags (``\textit{steer left}'', ``\textit{hard brake}'').  These outputs improve transparency, yet the continuous control signal is still handed off to conventional PID or MPC modules.  As datasets grew more diverse, two pain points surfaced.  First, the captioning interface incurred unacceptable latency when the vision backbone forwarded thousands of tokens.  Second, generic encoders wasted computation on pixels irrelevant to the driving task.  Proposals such as \textbf{TS-VLM}~\cite{chen2025tsvlm}, which introduced text-guided SoftSort pooling, and \textbf{DynRsl-VLM}~\cite{zhou2025dynrslvlm}, which adapted input resolution on demand, addressed these bottlenecks but did not yet close the semantic gap between textual narration and physical actuation.  Closing that gap became the next objective.

The earliest forays integrated language in a passive, descriptive role to enhance interpretability. A typical pipeline in this stage employed a frozen vision model (e.g. CLIP\cite{radford2021learning}) with an LLM decoder to explain the driving scene or recommended action in natural language, without directly outputting control. For example, DriveGPT-4 \cite{xu2024drivegpt4} would take a single front-camera image and produce either a textual description or a high-level maneuver label (“slow down”, “turn left”). These outputs helped explain what the perception system saw or intended, improving transparency. However, the actual vehicle control was still handled by conventional modules (PID controllers, etc.), so the language was an overlay rather than integral to decision-making. Moreover, two issues became apparent: (i) Generating long descriptions for each frame introduced latency, as vision encoders processed thousands of tokens per image\cite{zhou2025autovla}; (ii) General-purpose visual encoders wasted effort on irrelevant details, since not everything in an image is pertinent to driving\cite{zhai2025world}. Researchers responded with optimizations like TS-VLM \cite{chen2025tsvlm}, which uses text-guided soft attention pooling to focus on key regions, and DynRsl-VLM \cite{zhou2025dynrslvlm}, which dynamically adjusts input resolution to balance speed and detail. These improved efficiency, but there remained a semantic gap - narrating or labeling the scene is not the same as generating a precise steering or braking command. Closing that gap was the next logical step.

\newcommand{\cam}{\textsc{Cam}}
\newcommand{\lidar}{\textsc{L}}
\newcommand{\map}{\textsc{M}}
\newcommand{\instr}{I\textsubscript{in}}
\newcommand{\out}{I\textsubscript{out}}
\newcommand{\internal}{I\textsubscript{int}}
\newcommand{\llc}{\textbf{LLC}}
\newcommand{\traj}{\textbf{Traj.}}
\newcommand{\multi}{\textbf{Multi.}}

\begin{table*}[t]
  \centering
  \small
  \setlength{\tabcolsep}{2.7pt}
  % \caption{\textbf{Representative VLA4AD Models (2023–2025).}
  %          CAM = Multi-view camera input, L = LiDAR, M = HD-Map;
  %          Mono = single forward-facing camera;
  %          Outputs: \llc = low-level control, \traj = future trajectory. \multi= Multiple tasks such as perception, prediction or planning.}
  \caption{\textbf{Representative VLA4AD Models (2023--2025).}
           Sensor Inputs: Single = single forward-facing camera input;
           Multi = multi-view camera input;
           State = vehicle state information \& other sensor input.
           Outputs: \llc = low-level control, \traj = future trajectory, \multi = multiple tasks such as perception, prediction or planning.}
  \label{tab:vla4ad_models}
  \begin{tabularx}{\textwidth}{@{}l c c c c c c X@{}}
    \toprule
      &  & \multicolumn{2}{c}{\textbf{Data Source}} & \multicolumn{2}{c}{\textbf{Model}} &  &  \\
      \cmidrule(lr){3-4}\cmidrule(lr){5-6}
    \textbf{Model} & \textbf{Year} & \textbf{Input} & \textbf{Dataset / Benchmark} & \textbf{Vision} & \textbf{LLM} & \textbf{Output} & \textbf{Focus} \\
    \midrule
    DriveGPT-4~\cite{xu2024drivegpt4} & 2023 & Single  & BDD-X & CLIP & LLaMA-2 & \llc &
      \makecell[l]{Interpretable LLM\\Mixed Fine-tuning} \\[2pt]
    ADriver-I~\cite{jia2023adriveri} & 2023 & Single  & nuScenes + Private & CLIP ViT & Vicuna-1.5 & \llc &
      \makecell[l]{Diffusion World Model\\Vision–action Tokens} \\[2pt]   
    RAG-Driver~\cite{yuan2024rag} & 2024 & Multi  & BDD-X & CLIP ViT& Vicuna-1.5 & \llc &
      \makecell[l]{RAG Control\\Textual Rationales} \\[2pt]
      
    EMMA~\cite{hwang2024emma} & 2024 & Multi + State  & Waymo fleet & Gemini-VLM & Gemini & \multi &
      \makecell[l]{MLLM Backbone\\Multi-task Outputs} \\[2pt]
      
    CoVLA-Agent~\cite{chi2024covla} & 2024 & Single + State & CoVLA Data & CLIP ViT & Vicuna-1.5 & \traj &
      \makecell[l]{Text + Traj Outputs\\Auto-labelled Data} \\[2pt]
      
    OpenDriveVLA~\cite{zhou2025opendrivevla} & 2025 & Multi & nuScenes & Custom Module & Qwen-2.5 & \llc+\traj &
      \makecell[l]{2-D/3-D Align\\SOTA Planner} \\[2pt]

    ORION~\cite{xie2025orion} & 2025 & Multi + History & nuScenes + CARLA & QT-Former & Vicuna-1.5 & \traj &
      \makecell[l]{CoT Reasoning\\Continuous Actions} \\[2pt]

    DriveMoE~\cite{yang2025drivemoe} & 2025 & Multi& Bench2Drive & Paligemma-3B & – & \llc &
      \makecell[l]{Mixture-of-Experts\\Dynamic Routing} \\[2pt]
      
    VaViM~\cite{bartoccioni2025vavim} & 2025 & Video Frames & BDD100K + CARLA & LlamaGen & GPT-2 & \traj &
      \makecell[l]{Video-token PT\\Vision to Action} \\[2pt]
      
    DiffVLA~\cite{jiang2025diffvla} & 2025 & Multi + State & Navsim-v2 & CLIP ViT & Vicuna-1.5 & \traj &
      \makecell[l]{Mixed Diffusion\\VLM Sampling} \\[2pt]
      
    LangCoop~\cite{gao2025langcoop} & 2025 & Single + V2V & CARLA& GPT-4o & GPT-4o & \llc &
      \makecell[l]{Language-based V2V\\High Bandwidth Cut} \\[2pt]

     SimLingo~\cite{renz2025carllava} & 2025 & Multi & CARLA + Bench2Drive & InternVL2 & Qwen-2 & \llc+\traj &
      \makecell[l]{Enhanced VLM\\Action-dreaming} \\[2pt]
      
     SafeAuto~\cite{zhang2025safeauto} & 2025 & Multi + State  & BDD-X + DriveLM & CLIP ViT & Vicuna-1.5 & \llc &
      \makecell[l]{Traffic-Rule-Based\\PDCE Loss} \\[2pt]
      
    Impromptu-VLA~\cite{chi2025impromptu} & 2025 & Single & Impromptu Data & Qwen-2.5VL & Qwen-2.5VL & \traj &
      \makecell[l]{Corner-case QA\\NeuroNCAP SOTA} \\[2pt]
      
    AutoVLA~\cite{zhou2025autovla} & 2025 & Multi + State & nuScenes + CARLA & Qwen-2.5VL & Qwen-2.5VL & \llc+\traj &
      \makecell[l]{Adaptive Reasoning\\Multi Benchmark} \\[-2pt]
    \bottomrule
  \end{tabularx}
\end{table*}

\subsection{Modular VLA Models for AD}

As VLA research progressed, language evolved from a passive scene descriptor to an active planning component within modular architectures. Rather than merely commenting on the environment, language inputs and outputs began to inform planning decisions directly \cite{jiang2024senna}.
For example, OpenDriveVLA \cite{zhou2025opendrivevla} fused camera and LiDAR inputs with textual route instructions (e.g., “turn right at the church”), generating intermediate, human-readable waypoints (e.g., “turn right in 20m, then go straight”), which were then converted into continuous trajectories. This approach enhanced the transparency and flexibility of the planning process by introducing interpretable linguistic representations.

CoVLA-Agent \cite{chi2024covla} integrated visual and LiDAR tokens with optional textual prompts and used a compact MLP to map a selected action token (e.g., “turn left”) to a corresponding trajectory. Similarly, DriveMoE \cite{yang2025drivemoe} employed a Mixture-of-Experts architecture in which language cues were used to dynamically select sub-planners, such as an "overtaking expert" or "stop-and-go expert," based on the context.
In multi-agent scenarios, LangCoop \cite{gao2025langcoop} showed that vehicles could communicate using concise natural language messages to coordinate at intersections, representing a step toward language-enabled cooperation. SafeAuto \cite{zhang2025safeauto} incorporated symbolic traffic rules expressed in formal logic to validate or veto language-driven plans, ensuring that generated behaviors remained within safety constraints. Additionally, RAG-Driver \cite{yuan2024rag} introduced a retrieval-augmented planning mechanism, retrieving similar past driving cases from a memory bank to guide decision-making in ambiguous or long-tail scenarios.
Collectively, these approaches significantly reduced the semantic gap between language instructions and vehicle actions, effectively embedding natural language into the core of the planning loop. However, they often relied on multi-stage pipelines (perception → language → plan → control), which introduced latency and cascading failure risks at each module boundary. These limitations have motivated recent interest in more unified, end-to-end architectures, which aim to integrate perception, language understanding, and action generation within a single differentiable system.

\begin{table*}[t]
  \centering
  \small
  \setlength{\tabcolsep}{6pt}
  \caption{\textbf{Notable Datasets \& Benchmarks for VLA4AD.}
           Real = real-world logs; Sim = CARLA simulation\cite{dosovitskiy2017carla};
           clips = video segments; QA = question–answer pairs; CoT = chain-of-thought.}
  \label{tab:datasets_trimmed}
  \begin{tabularx}{\textwidth}{@{}l c c c c c@{}}
    \toprule
    \textbf{Name} & \textbf{Year} & \textbf{Domain} & \textbf{Scale} & \textbf{Modalities} & \textbf{Tasks} \\
    \midrule
    BDD100K / BDD-X~\cite{kim2018bddx,yu2018bdd100k}
      & 2018 & Real (US)        & 100\,k videos; 7\,k clips   & RGB video           & Captioning, QA    \\[4pt]
    nuScenes~\cite{caesar2020nuscenes}
      & 2020 & Real (Boston/SG)  & 1\,k scenes (20 s, 6 cams)  & RGB, LiDAR, Radar    & Detection, QA     \\[4pt]
    Bench2Drive~\cite{jia2024bench2drive}
      & 2024 & Sim (CARLA)       & 220 routes; 44 scenarios    & RGB                 & Closed-loop control \\[4pt]
    Reason2Drive~\cite{nie2024reason2drive}
      & 2024 & Real (nuSc/Waymo) & 600 k video–QA              & RGB video           & CoT-style QA      \\[4pt]
    DriveLM-Data~\cite{sima2024drivelm}
      & 2024 & Real+Sim          & 18 k scene graphs           & RGB, Graph          & Graph QA          \\[4pt]
    Impromptu VLA~\cite{chi2025impromptu}
      & 2025 & Real (multi-src)  & 80 k clips (30 s)           & RGB video, State    & QA, Trajectory    \\[4pt]
    NuInteract~\cite{zhao2025drivemonkey}
      & 2025 & Real (nuScenes)   & 1 k scenes                  & RGB, LiDAR          & Multi-turn QA     \\[4pt]
    DriveAction~\cite{hao2025driveaction}
      & 2025 & Real (fleet)      & 2.6 k scenarios; 16.2 k QA  & RGB video           & High-level QA     \\
    \bottomrule
  \end{tabularx}
\end{table*}

\subsection{Unified End-to-End VLA Models for AD}

With large multimodal foundation models available, researchers moved to fully unified networks that map sensors (and optional text commands) directly to trajectories or control signals in a single forward pass. A prime example is EMMA \cite{hwang2024emma}, which trains a massive VLM on Waymo’s autonomous driving data to jointly perform object detection and motion planning; the model learns a shared representation that serves both tasks, achieving better closed-loop performance than separate components. SimLingo\cite{renz2025simlingo}, LMDrive\cite{shao2024lmdrive} and CarLLaVA\cite{renz2025carllava} built on a LLaVA and fine-tuned it in CARLA simulator to follow language instructions and drive, notably introducing an “action dreaming” technique where the model imagines diverse outcomes for a given scenario by varying the language instruction, thus forcing a tight coupling between linguistic commands and the resulting trajectories. Other innovative approaches include using generative video models: ADriver-I \cite{jia2023adriveri} learned a latent world model that predicts future camera frames given actions (using diffusion), thereby enabling planning via imagining the consequences of actions. DiffVLA \cite{jiang2025diffvla} combined sparse (waypoints) and dense (occupancy grid) diffusion predictions to generate a trajectory conditioned on a textual scenario description, effectively sampling from a distribution of plausible safe maneuvers. End-to-end VLA models are highly reactive and effective at sensorimotor mapping, but a new limitation became clear: they can still struggle with long-horizon reasoning (planning far ahead or considering complex contingencies) and with providing fine-grained explanations of their decisions. 

% Interestingly, some work (e.g. VaVAM \cite{bartoccioni2025vavim}) finds that if a model is trained end-to-end on video inputs to output actions, it may implicitly learn “language-like” internal representations (e.g. identifying objects and intents) even without explicit language supervision, suggesting that much of the benefit of language might come from the structure it provides. 

\subsection{Reasoning-Augmented VLA Models for AD}
The latest wave of VLA4AD moves beyond explaning and plan conditioning toward \emph{long-horizon reasoning, memory, and interactivity}, placing VLMs/LLMs at the center of the control loop.  ORION \cite{fu2025orion} couples a transformer “QT-Former” memory, storing several minutes of observations and actions, with an LLM that
summarizes this history and outputs the next trajectory and a corresponding natural language explanation. Impromptu VLA \cite{chi2025impromptu} instead aligns CoT with action. Trained on 80k corner-case clips whose correct
reasoning steps are annotated, the model learns to verbalise its
decision path before acting, delivering state-of-the-art zero-shot negotiation between vehicles. AutoVLA \cite{zhou2025autovla} fuses CoT reasoning and trajectory planning within a single autoregressive transformer that tokenises continuous paths into discrete drive tokens, delivering state-of-the-art closed-loop success rates on nuPlan and CARLA. Collectively, these systems no longer just react to sensor input; they \emph{explain}, \emph{anticipate}, and carry out long-horizon \emph{reasoning} before outputing actions. They point toward conversational AVs that can verbally justify actions in real time, yet they surface new challenges: indexing city-scale memories, keeping LLM reasoning within a 30 Hz control loop, and formally verifying language-conditioned policies.

In summary, VLA4AD models have evolved from using language as a passive explanatory tool to integrating it as an active component in perception and control. We observe a steady closing of the loop between seeing, speaking, and acting—starting from explanatory perceptions, to modular VLA planning, to fully unified pipelines with reasoning and dialogue.  This progression points toward autonomous vehicles as conversational, collaborative agents—capable of not only safe driving, but also communication and reasoning aligned with human expectations.

\section{Datasets and Benchmarks}

We review several key datasets and evaluation suites, summarized in Table \ref{tab:datasets_trimmed}.

\paragraph{BDD\,100K / BDD-X~\cite{kim2018bddx,yu2018bdd100k}.}
BDD\,100K offers \textbf{100 k} diverse US videos; the BDD-X subset
(\(\sim\)7 k clips) adds time-aligned human \emph{rationales} (e.g., “slows because pedestrian crossing''), providing ground-truth explanations for models such as CoVLA-Agent \cite{chi2024covla} and SafeAuto \cite{zhang2025safeauto}.

\paragraph{nuScenes~\cite{caesar2020nuscenes}.}
\textbf{1k} 20s real-world episodes (Boston, Singapore) with 6 cams,
LiDAR + radar and full 3D labels.  Although language-free, it has been used for extensive VLA4AD evaluations.

\paragraph{Bench2Drive~\cite{jia2024bench2drive}.}
A closed-loop CARLA benchmark with \textbf{44} scenario types (\(220\) routes) and a 2M-frame training set. Metrics isolate specific skills (unprotected turns, cut-ins, etc.); DriveMoE \cite{yang2025drivemoe} tops the Leaderboard via specialized experts.

\paragraph{Reason2Drive~\cite{nie2024reason2drive}.}  
\textbf{600\,k} video–text pairs (nuScenes, Waymo, ONCE) annotated with CoT QA spanning perception \(\rightarrow\) prediction \(\rightarrow\) action. It evaluates logical consistency across entire reasoning chain using a \emph{consistency} metric to penalize incoherent multi-step answers.

\paragraph{DriveLM-Data~\cite{sima2024drivelm}.}
Provides graph-structured QA on nuScenes (\(18\) k graphs) and CARLA
(\(16\) k) scenes, stressing conditional reasoning. Baseline TS-VLM \cite{chen2025tsvlm} attains BLEU-4 56 but low graph consistency, leaving
ample room for improved multi-step reasoning.

\paragraph{Impromptu VLA~\cite{chi2025impromptu}.}
\textbf{80k} 30s clips (\(\sim\)2M frames) mined from eight public sets, curated for corner-case traffic (dense crowds, ambulances, adverse weather). Each clip pairs an expert trajectory and high-level instruction with rich captions and time-stamped QA. Provides an open evaluation server; training on this corpus yields measurable safety gains in closed-loop tests.

\paragraph{NuInteract~\cite{zhao2025drivemonkey}.}
Extends nuScenes with \textbf{1k} multi-view scenes that contain
dense captions and multi-turn 3D QA pairs, tightly linked to LiDAR ground truth. Supports multi-camera VQA and 3D reasoning; DriveMonkey shows substantial gains in cross-view QA when trained on this set.

\paragraph{DriveAction~\cite{hao2025driveaction}.}
A user-contributed, real-world benchmark containing \textbf{2.6 k} driving
scenarios and \textbf{16.2 k} vision-language QA pairs with \emph{action-level}
labels. The dataset spans broad, in-the-wild situations and offers
evaluation protocols that score VLA models on human-preferred driving
decisions, filling the gap left by perception-only suites.

In short, the datasets span the full spectrum needed for VLA4AD research: BDD-X \cite{kim2018bddx} and nuScenes \cite{caesar2020nuscenes} deliver large-scale, sensor-rich realism; Bench2Drive \cite{jia2024bench2drive} and Impromptu VLA \cite{chi2025impromptu} inject safety-critical corner cases; and Reason2Drive \cite{nie2024reason2drive}, DriveLM \cite{sima2024drivelm}, NuInteract \cite{zhao2025drivemonkey}, and DriveAction \cite{hao2025driveaction} supply the structured language needed for fine-grained reasoning and human-aligned actions. Harnessing these complementary assets is essential for training and benchmarking the next generation of VLA4AD.

\section{Training and Evaluation Strategies}
\label{sec:train_eval}

Building a VLA4AD policy involves two coupled goals: (i) learning a safe and competent
\emph{driving controller} and (ii) retaining a faithful \emph{language interface}.
Because driving data are costly and risky to collect, most work adopts a
\textit{pre-train \textrightarrow{} fine-tune} pipeline: behaviour cloning on large logs,
followed by targeted refinement in simulation or with rule–based constraints.
Below we review the two dominant paradigms—\textbf{Imitation Learning} and
\textbf{Reinforcement Learning}—and highlight how recent systems combine them.

\subsection{Training Paradigms}

\paragraph{Supervised Imitation Learning (IL).}
IL remains the work-horse for VLA4AD: the network ingests sensor streams
(and, if present, a language prompt) and minimises an
$\ell_2$ or cross-entropy loss to reproduce the expert’s control or trajectory.
CoVLA-Agent~\cite{chi2024covla} learns both a future path and a scene caption per frame,
while DriveMoE~\cite{yang2025drivemoe} and models trained on the SimLingo corpus
(CarLLAVA)~\cite{renz2025carllava} clone millions of simulator demonstrations.  
Although IL scales easily, it mirrors the training distribution; rare hazards
(e.g.\ cut-ins, occluded pedestrians) receive little supervision.
Typical remedies are \emph{DAgger-style} noisy roll-outs or explicit
corner-case augmentation, yet distribution drift can still cascade when
perception or language grounding fails.

\paragraph{Reinforcement Learning (RL).}
RL is usually layered on top of an IL warm-start.  
The policy interacts with a simulator (CARLA, Bench2Drive, etc.) and is
optimised with PPO or DQN style updates for route progress, collision
avoidance, and traffic-rule compliance.
Multi-agent settings such as LangCoop~\cite{gao2025langcoop} use RL to
refine V2V coordination, while SafeAuto~\cite{zhang2025safeauto} embeds
logical traffic rules as hard constraints or additional penalties.
A key open question is how to blend \emph{driving rewards} with
\emph{language fidelity}: current work often sidesteps the issue by freezing
the LLM and penalising only unsafe actions, leaving joint gradients over
text and control largely unexplored.  
As a result, RL for VLA is promising—especially for edge-case robustness—
but still under-developed compared with pure IL.

% \paragraph{Multi-stage \& Curricular Training.}
% Most VLA4AD systems adopt a phased recipe:
% \emph{(i) Foundation pre-training} on generic image/video corpora or internet
% text provides strong priors for the vision backbone and LLM;
% \emph{(ii) Task fine-tuning} then conditions the frozen or LoRA-adapted LLM
% on driving data that contain images, language, and actions—
% e.g.\ DriveMonkey is a LLaVA derivative fine-tuned on the
% NuInteract set to answer multi-view questions~\cite{zhao2025drivemonkey};
% \emph{(iii) Curriculum learning} feeds data from simple highway scenes to
% rare long-tail hazards.  SimLingo explicitly skips trivial
% cases so the policy focuses on instruction-conditioned
% corner events~\cite{renz2025carllava}, while
% Bench2Drive provides a simulator curriculum of 44 scenario
% types~\cite{jia2024bench2drive}.

\paragraph{Multi-stage Training.}  
Most VLA4AD models are trained via a four-stage curriculum:  
(1) \emph{Pre-train} large vision encoders (e.g., CLIP, InternViT) and language models (e.g., LLaMA, Vicuna) on broad image–text corpora and video datasets to learn general visual and linguistic priors;  
(2) \emph{Align} modalities by fine-tuning on paired image–text–action data—such as DriveMonkey on NuInteract \cite{zhao2025drivemonkey}—using cross-modal contrastive losses and sequence modeling objectives to bind scene features, language prompts, and control tokens;  
(3) \emph{Targeted augmentation} injects specialized traffic scenarios and instructions (e.g., SimLingo’s corner-case clips \cite{renz2025carllava}, Bench2Drive’s challenging routes \cite{jia2024bench2drive}), often supplemented with reinforcement learning or rule-based penalties (as in SafeAuto \cite{zhang2025safeauto}) to enforce safety constraints and improve performance on rare events;  
(4) \emph{Compress} the resulting model for deployment through parameter-efficient methods—LoRA adapters, sparse Mixture-of-Experts routing, or teacher–student distillation—reducing compute, memory, and latency while preserving the aligned VLA capabilities, as exemplified by DriveMoE \cite{yang2025drivemoe} and TS-VLM \cite{chen2025tsvlm}.

\paragraph{Balancing Language and Control.}
Joint losses typically weight a trajectory term against a
caption or QA term (e.g.\ CoVLA-Agent uses
$\mathcal{L}=\mathcal{L}_{\text{traj}}+\lambda\mathcal{L}_{\text{cap}}$).
Some authors alternate updates—one batch for driving, the
next for language—to avoid gradient interference.
Large LLMs are often kept frozen and prompted via a
light adapter; only the prompt-encoder is trained, preserving
linguistic fluency without huge GPU cost.
Free-form explanations complicate supervision:
CIDEr-style or RL-based caption tuning has been explored,
but care is required to reward factual accuracy over rhetoric.

\paragraph{Scalability and Efficiency.}
End-to-end VLA stacks (vision transformer + LLM + planner)
can exceed hundreds of GFLOPs per frame.
Current work therefore relies on:
\emph{LoRA and adapters} to update a few million parameters
inside a 70 B LLM (SafeAuto~\cite{zhang2025safeauto});
\emph{Mixture-of-Experts} routing so only a subset of specialists
run at inference (DriveMoE~\cite{yang2025drivemoe});
\emph{lightweight token reduction}—TS-VLM reports a $10\times$
speed-up by soft-attentive pooling~\cite{zhou2025dynrslvlm};
\emph{event-driven scheduling} that invokes the heavy model
only on scene changes; and 
\emph{distillation} (still rare in publications) to compress a
cluster-scale policy into an embedded “tiny VLA”.
A typical pipeline now freezes CLIP and LLaMA, trains a
small cross-attention head by imitation,
LoRA-adapts the LLM on language-augmented data,
optionally applies RL for red-light penalties,
and finally distils the result for on-car deployment.

% Overall, current practice follows a “clone → adapt → verify → compress’’
% loop, leaving open research on unified text+action RL and unsupervised
% hazard discovery.
% Together, these paradigms reveal a consensus workflow—\emph{clone, adapt,
% verify, compress}—while leaving open research space for unified
% text\,+\,action RL objectives and fully unsupervised corner-case
% discovery.

\subsection{Evaluation Protocols}
\label{subsec:evaluation}

Evaluating a VLA4AD agent is a \textit{dual-objective} task: the policy
must \emph{drive safely} \textbf{and} \emph{communicate faithfully}.
State-of-the-art papers therefore report four complementary metric
pillars:

\paragraph{Closed-loop Driving }
          \textit{Route success} on CARLA / Bench2Drive
          \cite{jia2024bench2drive}; \textit{infractions}
          (collisions, red-light, off-road) where DiffVLA halves error
          via a PDMS layer \cite{jiang2025diffvla};
          \textit{rule compliance} checked by logic vetoes
          (SafeAuto~\cite{zhang2025safeauto}); and
          \textit{generalisation} to unseen towns and weather.
          
\paragraph{Open-loop Prediction}
          \textit{Trajectory\,$\ell_2$} and collision rate (nuScenes
          challenge); \textit{goal reach} on instruction-conditioned
          targets; optional \textit{mAP/IoU} for auxiliary perception
          heads; and \textit{latency/FPS}—TS-VLM cuts compute by
          \(\sim 90\%\) through token pooling
          \cite{zhou2025dynrslvlm}.
          
\paragraph{Language Competence}
          \textit{Command follow} in SimLingo’s Action-Dreaming
          benchmark \cite{renz2025carllava}; automatic \textit{BLEU /
          CIDEr / accuracy} on NuInteract \cite{zhao2025drivemonkey} and
          DriveLM (BLEU-4 56 for TS-VLM \cite{sima2024drivelm});
          \textit{reason-chain consistency} in Reason2Drive
          \cite{nie2024reason2drive}; and \textit{human ratings} of
          BDD-X-style rationales.
          
\paragraph{Robustness \& Stress}
          \textit{Sensor perturbations} (blur, dropout, lag) analysed by
          DynRsl-VLM \cite{zhou2025dynrslvlm};
          \textit{adversarial prompts} or patches;
          \textit{out-of-distribution events}; and
          \textit{language edge-cases} (idioms, code-switching,
          multilingual queries).\\
          
Overall, a credible evaluation must measure
(i) control reliability, (ii) language fidelity, and (iii) their
coupling.  Today's suites cover these facets in isolation—CARLA /
Bench2Drive for control, NuInteract / Reason2Drive / DriveLM for
reasoning—highlighting the need for a unified \emph{“AI driver’s
licence''} that fuses both streams.
\section{Open Challenges}
\label{sec:challenges}

Despite rapid progress, Vision–Language–Action systems still face
substantial barriers before large–scale real–world deployment.
Below we summarise the six most pressing research fronts.

\paragraph{Robustness \& Reliability.}
Language reasoning adds context but opens new failure modes:
LLMs may hallucinate hazards or mis-parse slang (“floor it”).
Models must remain stable under sensor corruption
(rain, snow, glare) \emph{and} linguistic noise.
Logic–based safety vetoes such as SafeAuto~\cite{zhang2025safeauto}
are a first step, yet formal verification and “socially-compliant”
driving policies remain largely unexplored.

\paragraph{Real-time Performance.}
Running a vision transformer plus LLM at
\(\scriptstyle\ge 30\) Hz on automotive hardware is non-trivial.
Token–reduction designs like TS-VLM~\cite{zhou2025dynrslvlm},
hardware–aware quantisation, and event-triggered reasoning
(where heavy modules activate only on novel situations)
are promising; distillation or MoE sparsity will be required
as model sizes scale to billions of parameters.

\paragraph{Data \& Annotation Bottlenecks.}
Tri-modal supervision (image + control + language) is scarce and
costly—Impromptu VLA required 80k manually labelled clips.
Synthetic augmentation (e.g.\ SimLingo) helps,
but coverage of non-English dialects, traffic slang,
and legally binding phrasings is still thin.

\paragraph{Multimodal Alignment.}
Current VLA4AD work is camera-centric; LiDAR, radar,
HD-maps, and temporal state are only partially fused.
Approaches range from BEV projection of point-clouds to
3-D token adapters, and from ORION’s
language summarisation of long histories~\cite{xie2025orion}
to retrieval of textual map rules as in
RAG-Driver~\cite{yuan2024rag}.  A principled, temporally
consistent fusion of heterogeneous modalities is still missing.

\paragraph{Multi-agent Social Complexity.}
Scaling from pairwise coordination to dense traffic
raises protocol, trust, and security issues.
How should AVs exchange intent in a constrained yet flexible
“traffic language”?  Authentication and robustness to
malicious messages are open problems; cryptographic V2V and
gesture-to-text grounding are early research threads.

\paragraph{Domain Adaptation \& Evaluation.}
Sim-to-real transfer, cross-region generalisation, and
continual learning without catastrophic forgetting
are unresolved.  Community benchmarks (e.g.\ Bench2Drive)
cover only a fraction of the long-tail.
A regulatory “AI driver’s test'' that scores both control and
explanation quality is still to be defined.

In summary, addressing these challenges demands
joint advances in scalable training, formal safety analysis,
human–computer interaction, and policy.  Progress on any one
front—robust perception, efficient LLMs, trusted V2V, or
standardised evaluation—will accelerate the path toward
safe, transparent, and globally deployable VLA4AD systems.

\section{Future Directions}
\label{sec:future}

The next wave of research will likely widen the scope of VLA4AD from
prototype policies to \emph{scalable, cooperative, and verifiable} driving
platforms.  We highlight five promising threads.

\paragraph{Foundation–scale Driving Models.}
An obvious trajectory is a GPT-style ``driving backbone'':
a self-supervised, multi-sensor model trained on
dash-cams, LiDAR sweeps, HD-maps, and textual road rules.
Such a model could be prompted or LoRA-adapted for downstream
tasks with little data, similar to the way SimLingo/CarLLAVA
leverages instruction-conditioned trajectories~\cite{renz2025carllava}.
Realising this vision demands masked multimodal objectives and
architectures that process panoramic video together with free text.

\paragraph{Neuro-symbolic Safety Kernels.}
Pure end-to-end nets struggle to \emph{guarantee} safety.
Recent hybrids add rule layers—e.g.\ SafeAuto inserts logical
traffic checks~\cite{zhang2025safeauto}.  Future work may
let a neural VLA stack output a structured action program
(or CoT plan) that a symbolic verifier executes,
bridging flexibility and certifiability; ORION’s language
memory hints at such an interface~\cite{xie2025orion}.

\paragraph{Fleet-scale Continual Learning.}
Deployed AVs will encounter novel hazards daily.
Instead of raw logs, cars could uplink concise
language snippets (“new flagger pattern at $x,y$”) which are
aggregated into curriculum updates, much as
SimLingo filters trivial scenes to emphasise rare ones~\cite{renz2025carllava}.
Cloud agents could even answer real-time queries from
uncertain vehicles, boot-strapping knowledge across the fleet.

\paragraph{Standardised Traffic Language.}
Wide-area coordination will require a constrained,
ontology-driven message set—``I-yield-to-you'', ``Obstacle-ahead'',
etc.—analogous to ICAO phraseology in aviation.
VLA models are natural translators from raw perception to
such canonical intents; MoE routing (DriveMoE~\cite{yang2025drivemoe})
or token-reduction LMs (TS-VLM~\cite{zhou2025dynrslvlm})
can keep the bandwidth low enough for V2V links.

\paragraph{Cross-modal Social Intelligence.}
Future systems must parse gestures, voice, and signage as part
of the “language” channel—e.g.\ recognising a police
hand-signal or a pedestrian wave, then producing an explicit,
human-readable response (lights, display, honk).
Retrieval-augmented planners such as RAG-Driver~\cite{yuan2024rag}
suggest one route: fuse live perception with symbolic rules
and context to ground non-verbal cues.
Extending this to robust gesture–language–action alignment
remains open.

In essence, achieving these goals will require progress
in large-scale multimodal learning, formal verification,
communication standards, and human-AI interaction.
Success would yield a versatile “driving brain'' that can be
quickly adapted, safely audited, and seamlessly integrated
into the global traffic ecosystem.

\section{Conclusion}
In this work, we present the \textbf{first} comprehensive survey of Vision–Language–Action models for Autonomous Driving (VLA4AD), unifying several representative methods under a concise taxonomy that captures input modalities, core architectural components, and output formats. We trace the evolution of VLA4AD through four successive waves—\emph{Pre-VLA Explainers}, \emph{Modular VLA4AD}, \emph{End-to-End VLA4AD}, and \emph{Reasoning-Augmented VLA4AD}—highlighting how each stage has progressively closed the loop between perception, language understanding, and control.

We provide an in-depth comparison of training paradigms, from large-scale pre-training and modality alignment to targeted augmentation with corner-case data and efficient compression techniques, illustrating how these multi-stage workflows yield models that are both expressive and deployable. Our review of datasets and benchmark underscores the critical role of rich, multi-sensor, and language-grounded corpora in advancing VLA4AD capabilities.

Despite rapid progress, significant challenges remain: ensuring sub-30 Hz reasoning throughput, formal verification of language-conditioned policies, robust generalization in long-tail scenarios, and seamless sim-to-real transfer. We argue that the community must converge on shared evaluation protocols and open-source toolkits, invest in scalable memory and causal reasoning backbones, and pursue continual, fleet-scale learning to bridge research prototypes and production systems. By distilling current achievements and charting open avenues, we aim to inspire future work toward transparent, instruction-following, and socially aligned autonomous vehicles powered by integrated vision, language, and action.

\newpage
{\small
\bibliographystyle{ieee_fullname}
\bibliography{egbib}
}
\end{document}